# Fusion of Range and Thermal Images for Person Detection

W. Abbeloos[1], T. Goedemé[2]

[1] Electrical Engineering, KU Leuven, Sint-Katelijne-Waver, Belgium, wim.abbeloos@kuleuven.be
[2] Electrical Engineering, KU Leuven, Leuven, Belgium, tgoedeme@esat.kuleuven.be

*Abstract*— **Detecting people in images is a challenging problem. Differences in pose, clothing and lighting, along with other factors, cause a lot of variation in their appearance.**

**To overcome these issues, we propose a system based on fused range and thermal infrared images. These measurements show considerably less variation and provide more meaningful information. We provide a brief introduction to the sensor technology used and propose a calibration method. Several data fusion algorithms are compared and their performance is assessed on a simulated data set. The results of initial experiments on real data are analyzed and the measurement errors and the challenges they present are discussed.**

**The resulting fused data are used to efficiently detect people in a fixed camera set-up. The system is extended to include person tracking.**

## I. Introduction

Detecting and tracking people in images is an attractive method to monitor their movements. It is based on passive, non-contact sensors and hence does not disturb or distract the subjects. The analysis of the extracted position and pose data can be used in applications such as security and safety monitoring, home automation, patient monitoring or behavior analysis.

However, due to the large variation in peoples' appearance, it has proven to be a difficult problem to solve. Some solutions have been proposed but they typically have mediocre accuracy, suffer from severe limitations, require large amounts of annotated training data and are computationally expensive [1]. As an alternative, we investigate the use of other types of image sensors, which provide more informative input data. We use range and thermal image sensors, which provide measurements of physical properties much more stable and easier to interpret than the interaction of light and objects, as observed by a regular camera.

An important limitation of the thermal sensor used in our experiments is its low spatial resolution. To address this issue we discuss data fusion techniques allowing to upsample the temperature data using the higher resolution range data. While several papers suggest using high resolution sensor data to upsample range data [2, 3, 4], we are not aware of any work regarding the upsampling of other low resolution sensor data using range images. An automatic and convenient calibration method, which works despite the sensors low resolution, is proposed.

The performance of the upsampling algorithms is compared using an artificial dataset. We demonstrate the data fusion results on real data of some initial person detection experiments. Finally, we discuss some of the issues we experienced and the direction of our future work.

## II. Materials and methods

### A. Range camera

The range camera used in our system is a continuous wave time-of-flight (TOF) camera. The camera consists of two main components: the LED illumination array, which uses near infrared light (850nm) and is modulated at a high frequency (20 MHz), and the sensor, which is capable of measuring the phase of the reflected light. A phase measurement is available for every pixel and can be related directly to an absolute radial distance (e.g. in meters). The spatial resolution of our camera is 160 by 120 pixels.

If the intrinsic parameters of the camera are known, the distance measurement can be converted to 3D coordinates:

$$D_i \rightarrow (X, Y, Z)_i \qquad (1)$$

### B. Thermal camera

To acquire a thermal image we use a thermopile array. Every pixel of the array consists of a small thermocouple. The voltages generated by infrared (IR) waves emitted by objects in the observed scene are captured by the array and are digitized and read out. Compared to microbolometers, thermopile arrays are cheaper and smaller. The most important downside is the lower spatial resolution. Our sensor only has 16 by 16 pixels.

### C. Calibration

If we wish to combine the data of the thermal and range sensors, we must be able to directly relate every range measurement to its corresponding temperature measurement. If both sensors' intrinsic parameters are known (they can be determined using standard camera calibration techniques), this can be achieved by projecting the range measurement, with known (X,Y,Z) coordinates onto the thermal sensor using its known camera model. An overview is provided in figure 1.

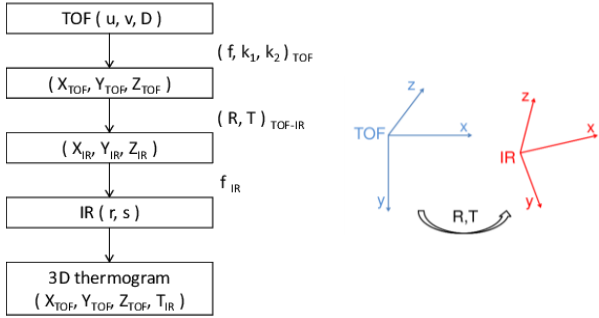

Fig. 1 Overall workflow of transforming and projecting the TOF pixels onto the IR array.

The only additional information required to be able to do this, is the relative pose of the sensors. While the translation can be measured with sufficient accuracy with a ruler or caliper, the rotation parameters must be determined from a series of calibration images.

While the calibration of two image sensors, e.g. in a stereo-camera setup is a solved problem, in this case we are faced with two problems:
1. The very low resolution of the thermopile array does not allow accurate estimation of the position of multiple points on a checkerboard pattern.
2. The two sensors measure two different physical properties. Our calibration target must have sufficient contrast in both sensors images.

To address these issues, we propose taking multiple calibration images of a single, circular calibration target. We use a heated copper plate, mounted at a certain distance in front of a planar (unheated) background. The target can then be segmented from the background and the center position easily determined using either the center of gravity, or by using the Hough circle detection algorithm [5].

The center points are then used to determine the 3 rotational parameters (Tait-Bryan angles) by minimizing the distance between their projection and the corresponding extracted point. To achieve the optimal parameters we minimize following error function:

$$E = \left| \mathbf{P}_{IR} - \mathbf{P} \times \mathbf{R} | \mathbf{T} \times \mathbf{P}_{TOF} \right| \quad (2)$$

Where $P_{TOF}$ contains the X,Y and Z coordinates of the point in the TOF cameras reference frame. The point is translated and rotated by $R$ and $T$. $T$ is the translation between the sensors. $R$ is the matrix below, containing the sines (s) and cosines (c) of the three Tait-Bryan angles.

$$R = \begin{bmatrix} c1 \times c3 & c1 \times s3 & -s1 \\ s2 \times s1 \times c3 - c2 \times s3 & s2 \times s1 \times s3 + c2 \times c3 & c1 \times s2 \\ c2 \times s1 \times s3 & c2 \times s1 \times s3 - s2 \times c3 & c1 \times c2 \end{bmatrix} \quad (3)$$

The points are then projected onto the IR pixel array using projection matrix $P$. To initialize, we set the parameters to zero.

*D. Data fusion and simulated evaluation dataset*

The goal is to associate a temperature with every range measurement. To do this we project the TOF pixel onto the IR array, as discussed before. A naïve approach may be to simply pick the projected points nearest neighbour on the thermal sensor.

We propose a way to utilize the range image to achieve superior upsampling accuracy. To measure and compare the performance of the algorithms, we generate an artificial scene in which we assign a temperature and depth to every object. We add noise and downsample the data to approximate the output of our real sensor system. The simulated dataset, including the temperature ground truth which will be used for comparison, is shown in figure 2. To assess the algorithm performance we use the absolute difference of the resulting temperatures of the algorithm output and the ground truth temperatures.

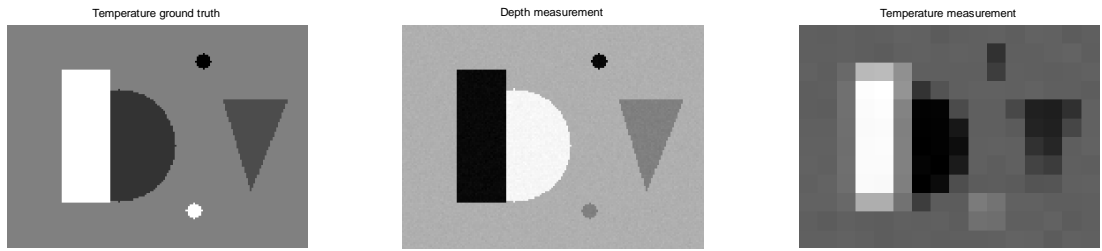

Fig. 2 An artificial scene was constructed to measure the upsampling accuracy. Every object was assigned a uniform temperature (left) and a depth (middle). Noise was added to both images. The high resolution thermal ground truth image was then downsampled to create a simulated low resolution thermal image (right), similar to what is measured by the sensors thermopile array.

*E. Proposed fusion algorithms*

In figure 3 we show the result of the naïve approach of Nearest Neighbour upsampling. We see that some significant errors occur near depth discontinuities. Using other standard upsampling techniques such as bilinear or bicubic interpolation, which use several points in the projections neighbourhood, results in the same type of errors.

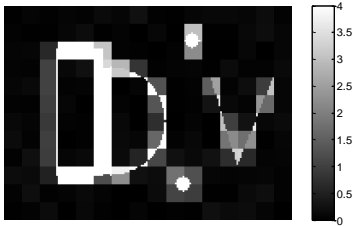

Fig. 3 Comparison of Neirest Neighbour upsampling to the ground truth temperature image (°C).

To increase the accuracy we propose new algorithms taking advantage of the available range measurement.

A first approach is to use a technique similar to bilinear interpolation but to use the depth-difference of the current TOF pixels projection to the mean depth of the four neighbouring IR pixels as the weights, instead of the relative position of its projection on the array (see figure 4).

$$W_i = \frac{|D - D_i|}{\sum_{i=1}^{4}|D - D_i|} \quad (4)$$

$$T = \mathbf{W} \times \mathbf{T} \quad (5)$$

The results are shown in figure 5. While this approach achieves some improvement near object boundaries, it still does not solve the averaging effect of the large IR pixels.

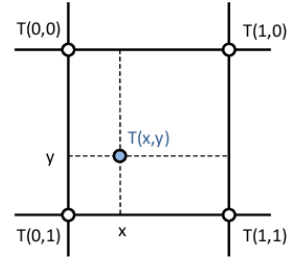

Fig. 4 The point *T(x,y)* is the projection of a TOF pixel onto the thermopile array. When using interpolation, the four neighbouring temperature measurements are used to assign a temperature to this point, depending on their relative position (indicated by x, and y).

To address this issue we propose a new algorithm that decomposes the pixel into multiple segments and estimates their individual temperature. The algorithm consists of following steps:

- Use k-means clustering on all depth values belonging to an IR pixel. This divides the pixel into different segments, each corresponding to an object.
- In the local 3x3 neighbourhood of the IR pixel, look for the neighbour with the best homogeneity (lowest depth difference between segments).
- Assign the temperature of this best neighbouring pixel to the segment of the current pixel that has the closest average depth.
- Compensate the temperature of the next segment using the known temperature of the first segment and the measured (average) temperature.

The temperature of second segment can be calculated by using following formula:

$$T_2 = \frac{T_m - T_1 \times A_1}{A_2} \quad (6)$$

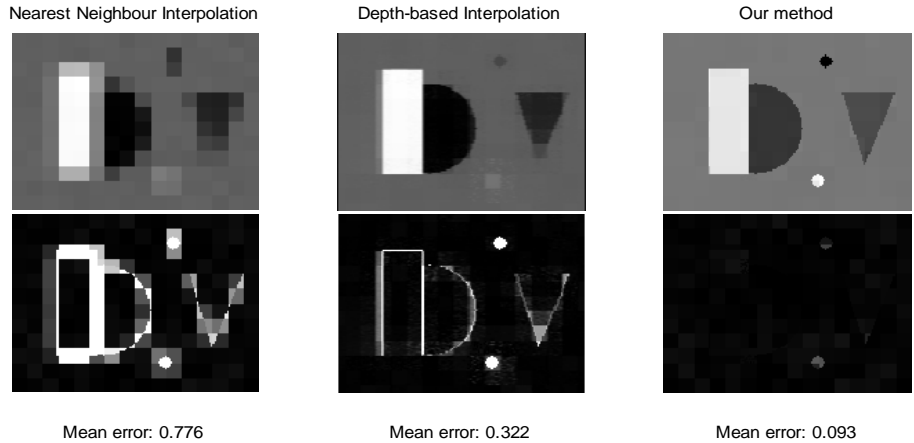

Nearest Neighbour Interpolation | Depth-based Interpolation | Our method

Mean error: 0.776    Mean error: 0.322    Mean error: 0.093

Fig. 3 The upper row shows the resulting upsampled thermal image, the lower row shows the absolute difference to the ground truth. The mean error per pixel (in °C) for the entire image is shown at the bottom. Our proposed method clearly outperforms methods that do not take advantage of the available range measurements. It also achieves higher accuracy then the simple depth-based interpolation method.

Where $T_1$, $T_2$ and $T_m$ are the temperatures of segment 1, segment 2 and the measured temperature, and $A_1$ and $A_2$ are the surface area (in pixels) of segment 1 and 2.

Note that this makes it possible to retrieve the correct temperature of a small object that is hotter or colder than its environment but only takes up a small part of the pixel.

The resulting upsampled thermal image is shown in figure 5 for visual comparison of the remaining errors. The error plot in figure 6 provides more detailed information on the occurring errors. We see that using our newly proposed algorithm, the mean error has been reduced to 1/8$^{th}$ compared to the naïve approach. Most of the remaining errors are due to the noise that was added.

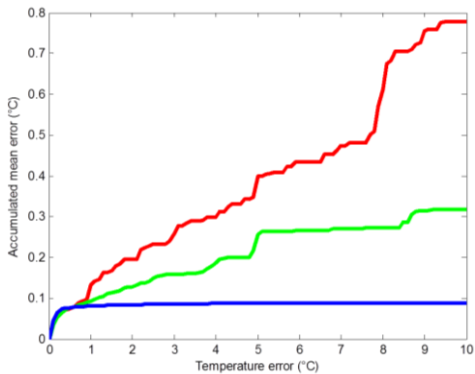

Figure 4: Accumulated error of the discussed up-sampling methods: Nearest Neighbour (red line), Depth-based (green line), and our method using area-based corrections (blue line).

### III. RESULTS AND DISCUSION

The developed data fusion techniques were implemented using the sensor system described in section 2. The algorithms were implemented on an embedded platform. The segmentation and temperature information can be combined with a tracking algorithm which allows monitoring the position and temperature of objects in time. As an example, we tracked a person walking through a room. This is illustrated in figure 7. The extracted mean temperature of the objects enables us to tell people from other moving objects such as chairs, tables, doors, and many more.

### IV. CONCLUSIONS

We have shown that by combining a low resolution thermopile array sensor with a higher resolution range camera, a lot of gain in resolution and accuracy of the measured temperatures can be achieved. The availability of temperature measurements greatly simplifies some problems that have proved very difficult to solve using regular cameras, such as the reliable detection of people.

In the future, we intend to combine the resulting improved thermal image with model-based detection [1] on regular, visual light images.

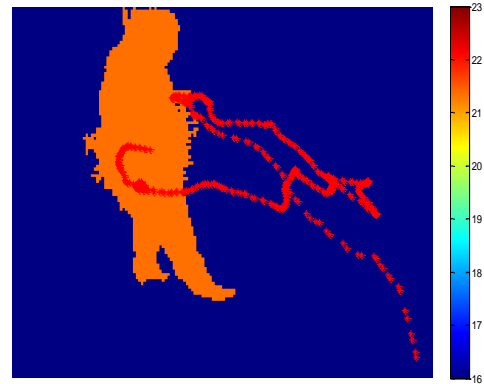

Fig. 5 Background subtraction allows detection of moving objects in the scene. The mean object temperature (in °C) is calculated and is used to tell people from other moving objects. Tracking the detected people (past positions indicated by the red points) allows to automatically extract a wide variety of statistics.